\documentclass[twoside,11pt]{article}

\usepackage[preprint]{jmlr2e}
\usepackage{booktabs}

\usepackage{xspace}
\usepackage{units}

\newcommand{\antmaze}{{\sc ant-maze}\xspace} 
\newcommand{\hcuni}{{\sc halfcheetah-uni}\xspace} 
\newcommand{\antuni}{{\sc ant-uni}\xspace} 
\newcommand{\walkeruni}{{\sc walker-uni}\xspace} 
\newcommand{\antomni}{{\sc ant-omni}\xspace}
\newcommand{\ptmaze}{{\sc point-maze}\xspace}
 
\newcommand{\anttrap}{{\sc ant-trap}\xspace}

\newcommand{\qdpg}{{\sc qd-pg}\xspace}

\newcommand{\mees}{{\sc me-es}\xspace}

\newcommand{\spea}{{\sc spea2}\xspace}
\newcommand{\nsga}{{\sc nsga-ii}\xspace}
\newcommand{\mome}{{\sc mome}\xspace}
\newcommand{\me}{{\sc map-e}lites\xspace}
\newcommand{\cvtme}{{\sc cvt map-e}lites\xspace}

\newcommand{\pgame}{{\sc pga-map-e}lites\xspace}

\newcommand{\cmame}{{\sc cma-me}\xspace}
\newcommand{\diayn}{{\sc diayn}\xspace}
\newcommand{\dads}{{\sc dads}\xspace}
\newcommand{\smerl}{{\sc smerl}\xspace}

\newcommand{\qd}{{\sc qd}\xspace}
\newcommand{\rl}{{\sc rl}\xspace}

\newcommand{\omgmega}{{\sc omg-mega}\xspace}
\newcommand{\cmamega}{{\sc cma-mega}\xspace}

\newcommand{\pbt}{{\sc pbt}\xspace}
\newcommand{\mepbt}{{\sc me-pbt}\xspace}

\newcommand{\qdrl}{{\sc qd-rl}\xspace}

\newcommand{\jax}{{\sc jax}\xspace}
\newcommand{\brax}{{\sc brax}\xspace}
\newcommand{\qdax}{{\sc qd}ax\xspace}

\newcommand{\sferes}{{\sc Sferes}\xspace}
\newcommand{\pyribs}{{\sc Pyribs}\xspace}
\newcommand{\qdpy}{{\sc qd}py\xspace}
\newcommand{\evojax}{{\sc E}vo{\sc jax}\xspace}
\newcommand{\evosax}{evosax\xspace}
\newcommand{\evotorch}{{\sc E}vo{\sc T}orch\xspace}

\usepackage{microtype}
\usepackage[skip=-1pt]{caption}

\jmlrheading{1}{2000}{1-48}{4/00}{10/00}{hello}{Chalumeau and Lim et al.}

\ShortHeadings{QDax: A Library for Quality-Diversity and Population-based Algorithms}{Chalumeau and Lim et al.}
\firstpageno{1}

\begin{document}

\title{QDax: A Library for Quality-Diversity and Population-based Algorithms with Hardware Acceleration}

\author{\name Felix Chalumeau* $^1$ \email f.chalumeau@instadeep.com \\
       \name Bryan Lim* $^2$ \email bryan.lim16@imperial.ac.uk \\
       \name Raphaël Boige \email r.boige@instadeep.com \\
       \name Maxime Allard \email m.allard20@imperial.ac.uk \\
       \name Luca Grillotti \email luca.grillotti16@imperial.ac.uk \\
       \name Manon Flageat \email manon.flageat18@imperial.ac.uk \\
       \name Valentin Macé \email v.mace@instadeep.com \\
       \name Guillaume Richard \email g.richard@instadeep.com \\
       \name Arthur Flajolet \email a.flajolet@instadeep.com \\
       \name Thomas Pierrot** $^1$ \email t.pierrot@instadeep.com \\
       \name Antoine Cully** $^2$ \email a.cully@imperial.ac.uk \\
       \AND
       \addr $^1$InstaDeep \quad
       \addr $^2$Department of Computing, Imperial College London \\
       \addr * Equal Contribution \quad
       \addr ** Equal Supervision
}

\editor{xxx}

\maketitle

\begin{abstract}
\qdax is an open-source library with a streamlined and modular API for Quality-Diversity (QD) optimization algorithms in \jax. The library serves as a versatile tool for optimization purposes, ranging from black-box optimization to continuous control. \qdax offers implementations of popular QD, Neuroevolution, and Reinforcement Learning (RL) algorithms, supported by various examples. All the implementations can be just-in-time compiled with Jax, facilitating efficient execution across multiple accelerators, including GPUs and TPUs. These implementations effectively demonstrate the framework's flexibility and user-friendliness, easing experimentation for research purposes. Furthermore, the library is thoroughly documented and tested with 95\% coverage.

\end{abstract}



\section{Introduction}
Quality Diversity (\qd) has emerged as a rapidly expanding family of stochastic optimization algorithms that have demonstrated competitive performance across diverse applications, including robotics~\citep{cully2015robots}, engineering design~\citep{gaier2017data}, and video games~\citep{alvarez2019empoweringqd}. Unlike traditional optimization algorithms that seek a single objective-maximizing solution, \qd algorithms aim to identify a diverse set of such solutions. Recently, the integration of modern deep reinforcement learning techniques with \qd~\citep{nilsson2021policy, pierrot2021diversity} has enabled the tackling of complex problems, such as high-dimensional control and decision-making tasks. This convergence has drawn parallels with other fields that maintain populations of policies and explore diverse behaviors~\citep{eysenbach2018diversity, sharma2020dads}. Overall, these algorithmic developments have gained popularity due to their simplicity, versatility, and ability to generate practical solutions for various industrial applications.

To facilitate the unification of algorithms from these distinct research areas, we propose a comprehensive framework embodied in a new open-source library, \qdax. \qdax encompasses meticulously crafted implementations of 16 methods, prioritizing both speed and flexibility. In particular, \qdax offers an extensive range of \qd algorithm implementations, all within a cohesive framework. Furthermore, \qdax provides Reinforcement Learning (\rl) algorithms and skill-discovery \rl algorithms, both built on the same foundational components. This commonality ensures reliability in comparative analysis and facilitates seamless integration of diverse approaches. Additionally, \qdax offers a collection of utilities tailored for benchmarking algorithms on standard tasks, including mathematical functions, robot control scenarios, and industry-driven decision-making problems.

\qdax is built on top of \jax~\citep{jax2018github}, a Python library for high-performance numerical computing and machine learning, to fully exploit modern hardware accelerators like GPUs and TPUs with minimal engineering overhead. \jax enables to harness the scalability advantages of \qd methods, as demonstrated in recent studies~\citep{lim2022accelerated}, and the parallelization capabilities of fast parallel simulators, such as those developed in \brax~\citep{brax2021github} and Isaac gym~\citep{makoviychuk2021isaac}. Because it provides reliable implementations and faster benchmarking methods, \qdax represents a significant stride towards accelerating the development of QD and population-based methods.


\begin{figure*}[t]
    \centering
    \includegraphics[width=.85\textwidth]{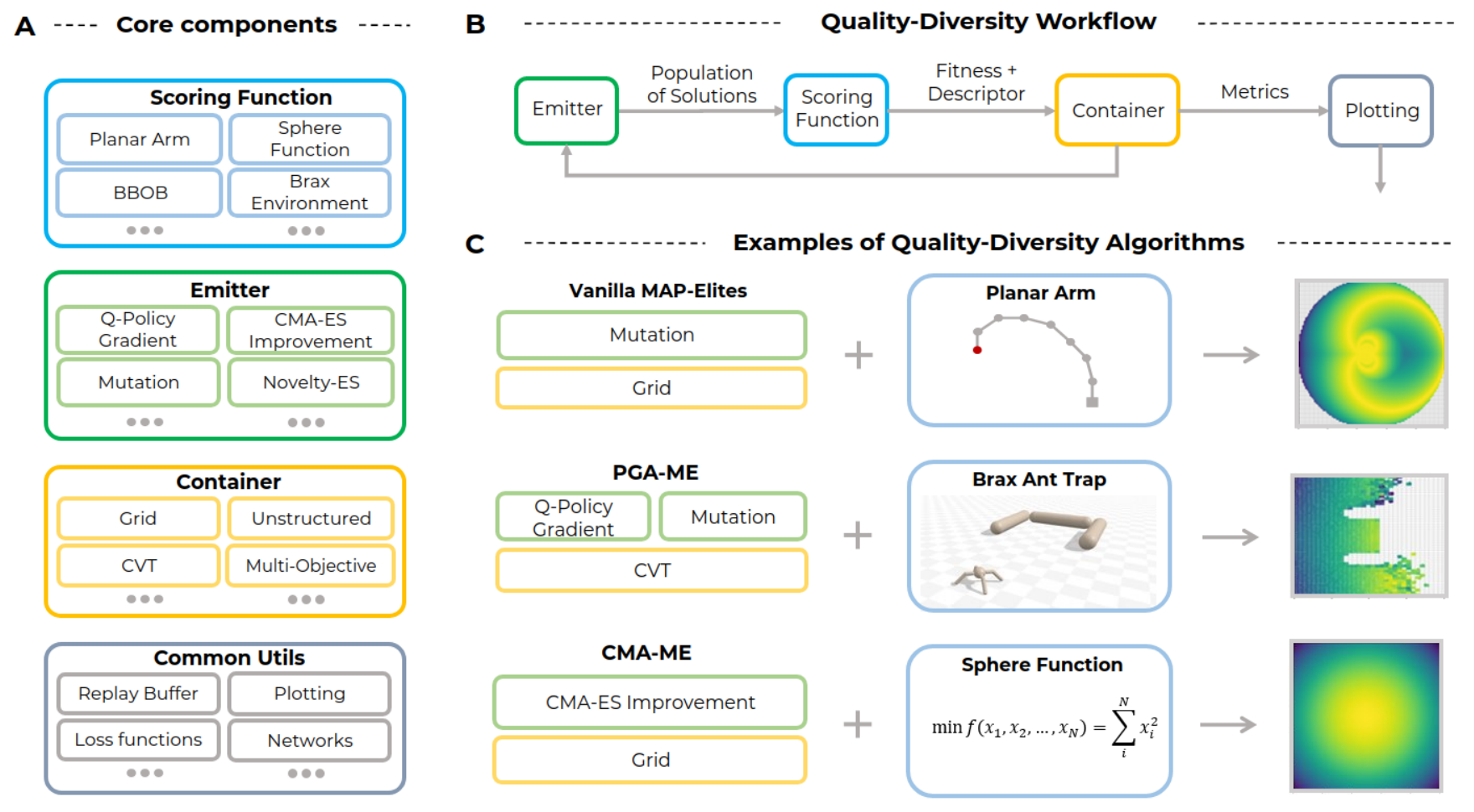}
    \caption{\textbf{A}: Core components, used as building blocks to create optimization experiments. \textbf{B}: High-level software architecture of QDax. \textbf{C}: Various examples of QD algorithms used for a variety of tasks and problem settings available in \qdax.}
    \label{fig:architecture}
    \vspace*{-3mm}
\end{figure*}

\section{QDax Features}

\paragraph{Comprehensive baselines and SOTA methods} \qdax includes numerous baselines for \qd, Skill-Discovery \rl, Population-Based Training, including state-of-the-art methods from those fields. Implementations' performance are validated through replications of results from the literature. Methods can be run in only a few lines of codes thanks to a simple API. Additionally, the flexibility of this API enables to extend those methods for research purposes. See Appendix~\ref{app:api} for examples of API use.


\paragraph{Fast runtimes.} Algorithms and tasks are implemented in \jax and are fully compatible with just-in-time (jit) compilation to take full advantage of XLA optimization and to be able to seamlessly leverage hardware accelerators (e.g. GPUs and TPUs). Practitioners can expect to speed up their runtimes by up to 10x compared to existing implementations, thereby reducing experiment times from hours to mere minutes (see Appendix~\ref{app:fast}).

\paragraph{Optimization problems.}
To enhance reproducibility, \qdax offers \jax implementations of widely studied problems. \qdax includes a collection of fundamental mathematical functions that serve as elementary benchmarks in \qd, such as rastrigin, sphere, planar arm, and various others. To further expand the repertoire of tasks, \qdax provides a range of utilities for deriving tasks from robotics simulation environments implemented in \brax. These utilities incorporate robotics motion \qd tasks, as inspired by \citet{flageat2022benchmarking}, and hard exploration tasks, as presented in \citet{chalumeau2022assessing}. Additionally, we provide support for the RL industrial environments suite Jumanji \citep{jumanji2022github}, facilitating the evaluation on Combinatorial Optimization problems. For a comprehensive analysis of the performance of various \qdax algorithms on selected tasks, refer to Appendix~\ref{app:results}.

\paragraph{Documentation, Examples and Code.} \qdax provides extensive documentation of the entire library\footnote{\url{https://qdax.readthedocs.io/en/latest}/}. All classes and functions are typed and described with docstrings. Most functions are covered by unit tests (95\% code coverage), with publicly available reports\footnote{\url{https://app.codecov.io/gh/adaptive-intelligent-robotics/QDax}}. Stability of the library is also ensured through continuous integration, which validates any change through type checking, style checking, unit tests and documentation building. \qdax also includes tutorial-style interactive Colab notebooks\footnote{\url{https://github.com/adaptive-intelligent-robotics/QDax/tree/main/examples}} which demonstrate example usage of the library through the browser without any prior setup. These examples help users get started with the library and also showcase advanced usage.
Docker and Singularity container functionalities are also provided to ease reproducibility and deployment on cloud servers. Conda support is also provided and can be used as an alternative. Finally, \qdax can be installed via PyPI\footnote{\url{https://pypi.org/project/qdax/}}.

\section{Architecture and Design of QDax}
\qdax is built on Python and \jax~\citep{jax2018github}. It can seamlessly run on CPUs, GPUs or TPUs on a single machine or in a distributed setting.
\qdax introduces a framework (see Figure~\ref{fig:architecture}) that unifies all the recent state-of-the-art \qd algorithms. In this framework, a \qd algorithm is defined by a \textbf{container} and an \textbf{emitter}. The \textbf{container} defines the way the population is stored and updated at each evolution step. The \textbf{emitter} implements the way solutions from the population are updated; this encompasses mutation-based updates, sampling from a distribution (Evolution Strategies), policy-gradient updates, and more. 
Neuroevolution methods - and more generally algorithms not rooted in QD - also fit in the framework, see Table~\ref{table:qdaxalgorithms} that highlights components that are shared across algorithms implemented in \qdax.

\qdax was also designed to compare and combine \qd algorithms with other algorithms, such as deep \rl and other population-based approaches. Consequently, the architecture ensures common use of many components, such as loss functions, networks and replay buffers. This helps building fair performance comparisons between several classes of algorithms and reduces the time it takes to implement hybrid approaches. The \textbf{scoring function} abstraction gives the user flexibility to define the optimization problem. It can range from standard optimization functions to complex rollouts in a simulated environment for \rl.

\begin{table}[t]
    \begin{center}
    \resizebox{0.85\linewidth}{!}{
    \begin{tabular}{lccc}
        \toprule
        \textbf{Algorithm} & \textbf{Container} & \textbf{Emitter} & \textbf{Common utils} \\
        \midrule
        \me & Grid & GA & Network\\
        \cmame & Grid & CMA-ES Impr. & Network, CMAES opt.\\
        \mees & Grid & ES & Network\\
        \pgame & CVT & GA, Q-PG & Network, Buffer, Loss\\
        \qdpg & CVT & GA, Q-PG, D-PG & Network, Buffer, Loss\\
        \omgmega & Grid & OMG-MEGA & N/A\\
        \cmamega & Grid & CMA-MEGA & CMAES opt.\\
        \mome &  Multi Objective Grid & GA & Pareto front\\
        \nsga, \spea &  Pareto Grid & GA & Pareto front\\
        \diayn, \dads, \smerl, \pbt & N/A & N/A & Network, Buffer, Loss\\
        \mepbt & Grid & GA, PBT & Network, Buffer, Loss\\
        \bottomrule
        
    \end{tabular}
    }
    \end{center}
    \caption{\textbf{Comprehensive baselines} implemented in \qdax, including state-of-the-art methods for \qd, Skill-Discovery and Multi-Objective Optimization.}
    \label{table:qdaxalgorithms}
    \vspace{-5mm}
\end{table}

\section{Comparison to existing libraries}
\vspace{-2mm}

Multiple independent open-source packages are currently accessible for quality diversity (\qd). One such framework is \sferes \citep{Mouret2010}, which offers a selection of \qd algorithms implemented in the C++ language. However, it lacks support and integration for deep learning frameworks. On the other hand, python libraries such as \pyribs~\citep{pyribs} and \qdpy~\citep{qdpy} provide a more extensive range of \qd algorithms compared to \sferes. While these libraries are user-friendly, they lack GPU or TPU acceleration and distribution capabilities. Furthermore, they are limited to standard optimization problems and do not support sequential decision-making problems. 

Additionally, recent developments in the field of evolutionary algorithms have resulted in the creation of several packages. For instance, \evojax~\citep{evojax2022} and \evosax~\citep{evosax2022github} are two \jax packages that offer efficient implementations of Evolution Strategies and Neuroevolution methods, as well as the implementation of optimization problems using \jax. These libraries do not focus on \qd and hence have a very limited number of \qd methods. Interestingly, these libraries can be used in conjunction with \qdax. Similarly, \evotorch builds upon PyTorch~\citep{NEURIPS2019_9015} rather than \jax to accelerate Evolutionary Algorithms.

Overall, \qdax presents a unified framework in \jax, facilitating rapid development and effortless benchmarking. It serves as a platform for evolutionary, population-based, and diversity-seeking algorithms, while leveraging modern hardware accelerators.




\acks{We would like to thank Rémi Debette and Achraf Garai for their help in this project. This research was supported with Cloud TPUs from Google's TPU Research Cloud (TRC).}


\noindent

\vskip 0.2in
\bibliography{main}

\newpage

\appendix



\section{Code API and Usage} \label{app:api}
\begin{figure*}[ht!]
    \centering
    \includegraphics[width=1.0\textwidth]{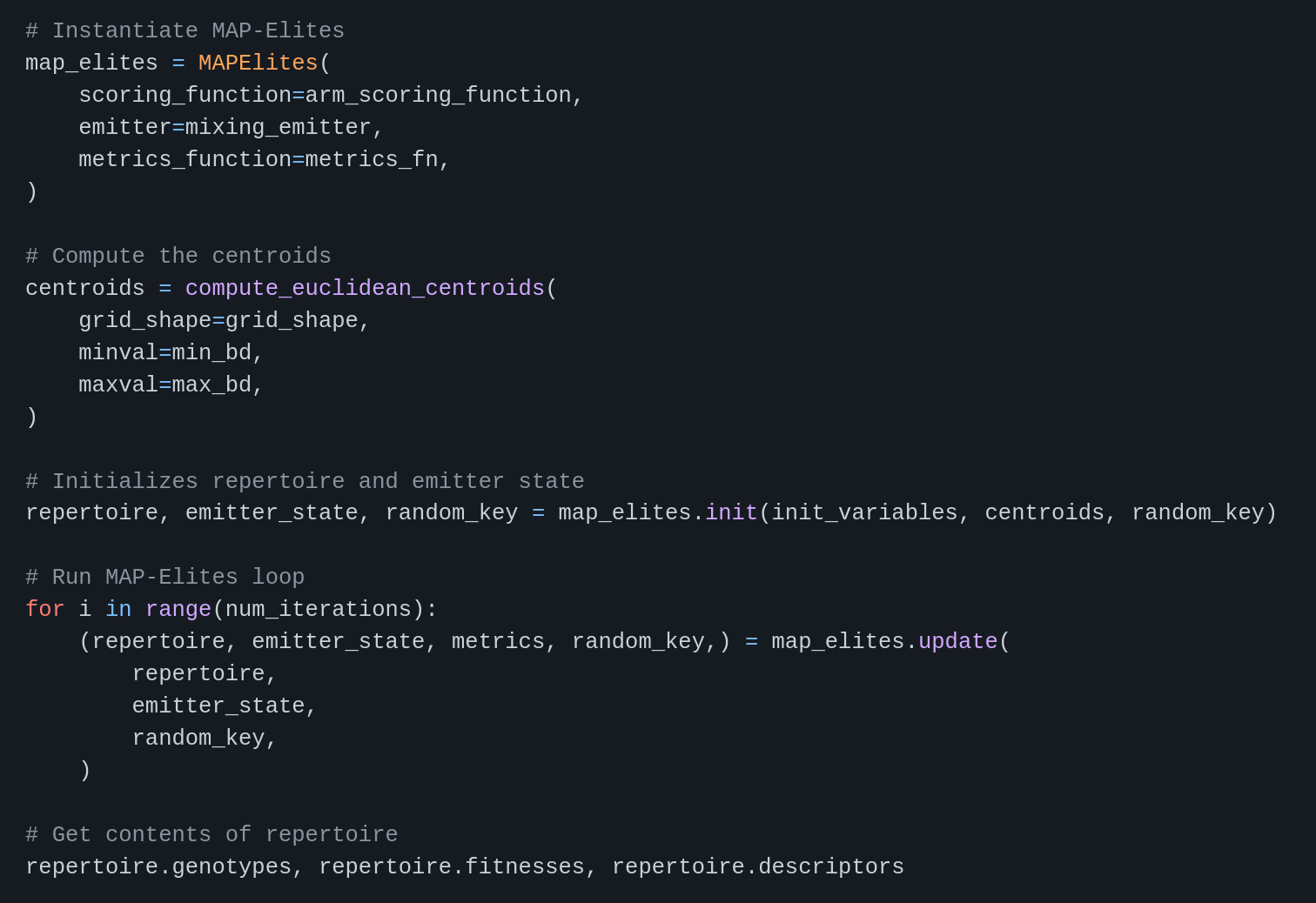}
    \caption{Code snippet demonstrating the API of \qdax on a simple example: \me used to solve the arm task. User can get their first experiment running with only a few lines of code. Additionally, it only takes a few updates to change the type of algorithm running on the same task.}
    \label{fig:apisnippet}
    \vspace*{-0.3cm}
\end{figure*}

Figure~\ref{fig:apisnippet} shows an example of the code usage demonstrating the simplicity and modularity of the API. This code snippet shows the main steps to run \me with \qdax, those steps being (i) instantiating an object from the class \me (ii) computing the centroids that will be used to define the grid that is to store the solutions produced by the algorithm (iii) initializing the algorithm, which will initialize the repertoire as well as the state of the emitter (iv) running iterations of the update function of \me, natively implemented in the class. Interestingly, all it takes to run \pgame or \cmame is to change the type of emitter defined when instantiating the \me object on the first line of the code snippet.

For additional illustrations of the API, we strongly encourage the reader to try the examples proposed in \qdax: we wrote example notebooks\footnote{\url{https://github.com/adaptive-intelligent-robotics/QDax/tree/main/examples}} for most algorithms implemented. This should help any user to get started with the library and should also illustrate how to use the library to extend the implemented algorithms for research purpose.

\section{Implemented algorithms} \label{app:algs}
The main algorithms implemented in \qdax are summarized in table~\ref{table:qdaxalgorithms_2}. This table provides the name of the algorithm, the broader categories of algorithms that the algorithm belongs to and the reference of the algorithm.

\begin{table} 
    \begin{center}
    \begin{tabular}{lcc}
        \toprule
        \textbf{Algorithm} & \textbf{Categories} & \textbf{References} \\
        \midrule
        \me &  \qd & \citet{mouret2015illuminating}\\
        \cvtme &  \qd & \citet{vassiliades2017using}\\
        \midrule
        \cmame &  \qd, Evolution Strategies & \citet{fontaine2020covariance}\\
        \mees &  \qd, Evolution Strategies & \citet{colas2020scaling}\\
        \midrule
        \pgame &  \qd, \rl & \citet{nilsson2021policy}\\
        \qdpg &  \qd, \rl & \citet{pierrot2021diversity}\\
        \midrule
        \diayn &  Skill Discovery, \rl & \citet{eysenbach2018diversity}\\
        \dads &  Skill Discovery, \rl & \citet{sharma2020dads}\\
        \smerl &  Skill Discovery, \rl & \citet{kumar2020smerl}\\
        \midrule
        \omgmega &  Differentiable \qd & \citet{fontaine2021differentiable}\\
        \cmamega &  Differentiable \qd & \citet{fontaine2021differentiable}\\
        \midrule
        \mome &  Multi Objective, \qd & \citet{pierrot2022multi}\\
        \nsga &  Multi Objective & \citet{deb2002nsga2}\\
        \spea &  Multi Objective & \citet{zitzler2001spea2}\\
        \midrule
        \pbt & Population Based, RL & \citet{jaderberg2017pbt}\\
        \mepbt & Population Based, QD, RL & \citet{pierrot2023evolving}\\
        \bottomrule

    \end{tabular}
    \end{center}
    \caption{Main algorithms implemented in \qdax. This table also reports the general category of algorithms each method belongs to, as well as the paper that introduced the algorithm.}
    \label{table:qdaxalgorithms_2}
\end{table}

\qdax contains state-of-the-art \qd algorithms and numerous methods related to Neuroevolution, Skill Discovery, Population-Based learning and Multi-Objective optimization. 
In particular, \qdax provides the implementation of \me~\citep{mouret2015illuminating}, \cvtme~\citep{vassiliades2017using} - corresponding to general \qd methods -, \cmame~\citep{fontaine2020covariance}, \mees~\citep{colas2020scaling} - described as \qd with Evolution Strategies - , \pgame~\citep{nilsson2021policy, flageat2022empirical}, \qdpg~\citep{pierrot2021diversity} - which are \qd methods with policy gradients, often referred to as \qdrl - , \diayn~\citep{eysenbach2018diversity}, \dads~\citep{sharma2020dads}, \smerl~\citep{kumar2020smerl} - popular methods in \rl for Skill Discovery -, \omgmega, \cmamega~\citep{fontaine2021differentiable} - Differentiable \qd - , \mome~\citep{pierrot2022multi}, \nsga, \spea - the reference approaches for Multi-Objective optimization -. All those methods have similar API and can be evaluated on popular tasks. Furthermore, they all take advantage from the speed-up enabled by just-in-time compilation in Jax, making them extremely fast and scalable.

\section{Benchmark Results} \label{app:results}
This section provides numerous results from the algorithms present in \qdax over several benchmarks tasks available in the library.  These results can give the reader an idea of the metrics and time performance expected when using \qdax. Note that metrics performance are validated against the one reported along original implementations of the algorithms. The reported experiments are taken from \citet{chaluboige2022neuroevolution} and were run with a single Quadro RTX 4000 GPU.

\begin{figure} 
    \centering
    \includegraphics[width=0.9\textwidth]{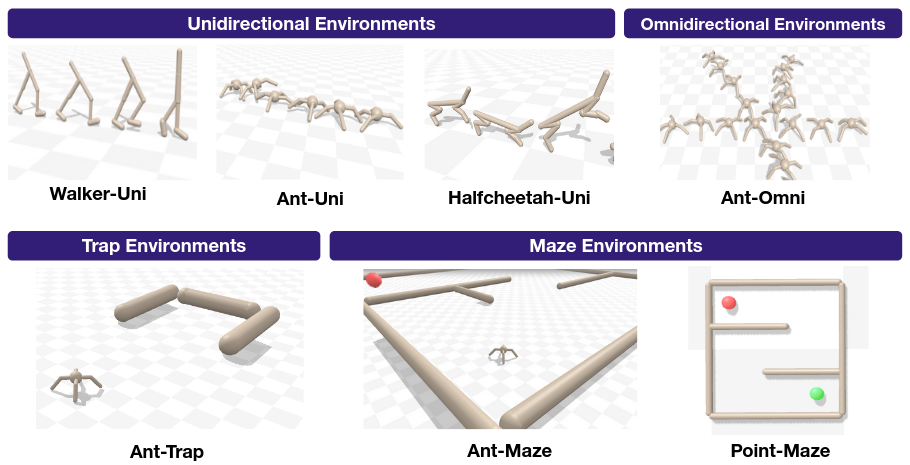}
    \caption{Visualisation of the environments used to show the performances of our implementations of \smerl, \pgame and \me. Diagram is adapted from \citet{chaluboige2022neuroevolution, flageat2022benchmarking, chalumeau2022assessing}}
    \label{fig:env_viz}
\end{figure}

The results presented relate to seven tasks from the \brax \rl tasks implemented in \qdax. These tasks are benchmark \qdrl tasks used to assess Neuroevolution algorithms~\citep{chalumeau2022assessing, flageat2022benchmarking}. We refer to them as \antuni, \hcuni, \walkeruni, \anttrap, \antmaze and \ptmaze. Those can be visualized on \autoref{fig:env_viz}. For more details about the hyper-parameters used, please refer to~\citep{chaluboige2022neuroevolution}. 

For a representative comparison between \qd algorithms and Deep \rl methods such as \smerl, which do not actively have a population of policies or an archive, a passive archive is used to compute metrics like \qd score in those results. Given that there is only a single latent conditioned policy, during training SMERL policy is evaluated by sampling latent codes and recording their trajectories. Behavior descriptors can be extracted from these trajectories, which can then be used for addition to the passive archive. This allows to use similar metrics such as the coverage and \qd score when comparing \qd and Skill Discovery \rl methods.

\begin{figure} 
    \centering
    \includegraphics[width=0.9\textwidth]{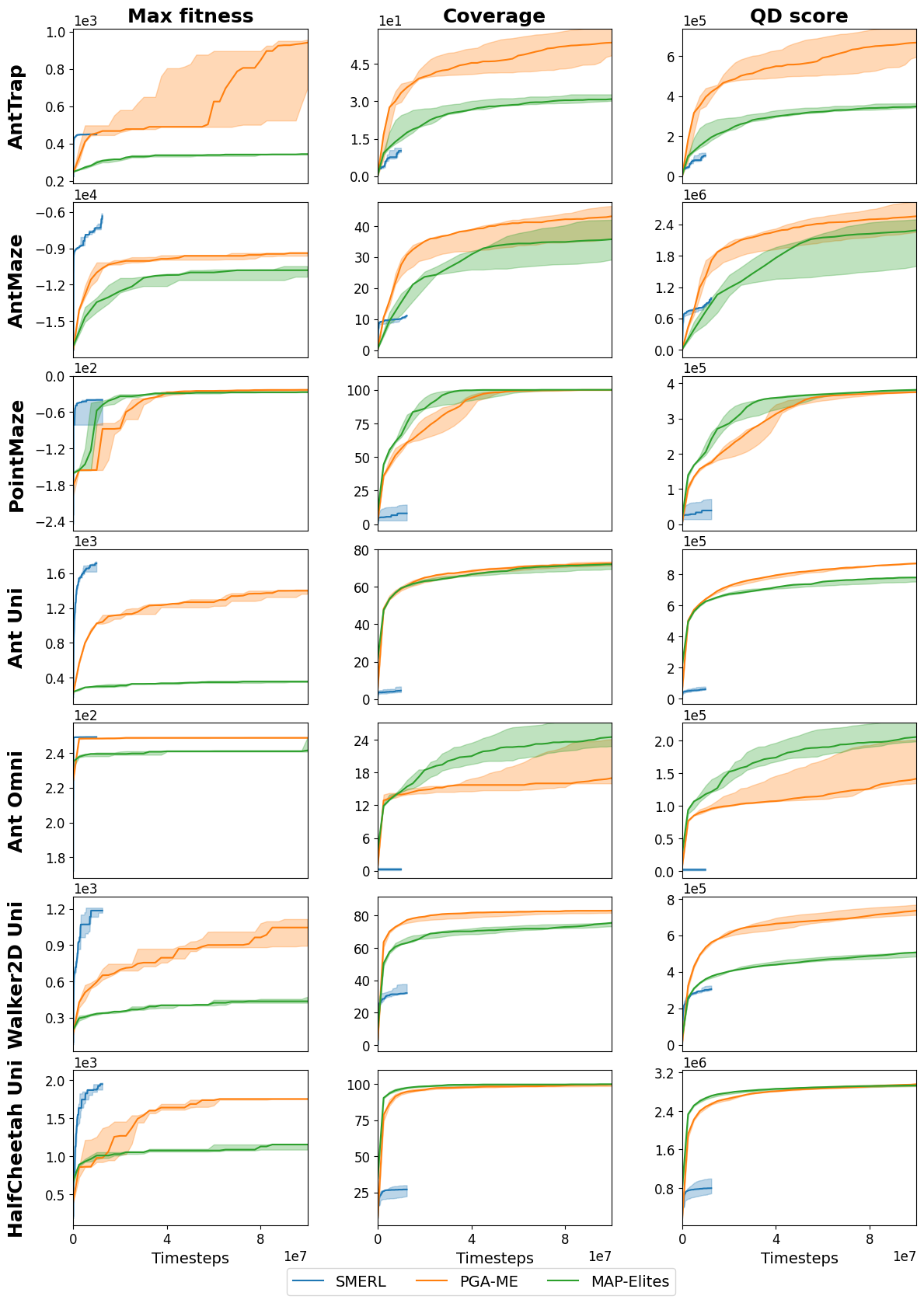}
    \caption{Evolution of the maximum fitness, coverage and \qd score along environment interactions, during a training phase. Reports \smerl on $10^7$ timesteps and \pgame and \me on $10^8$ timesteps.}
    \label{fig:qd_metrics_step}
\end{figure}

\begin{figure} 
    \centering
    \includegraphics[width=0.8\textwidth]{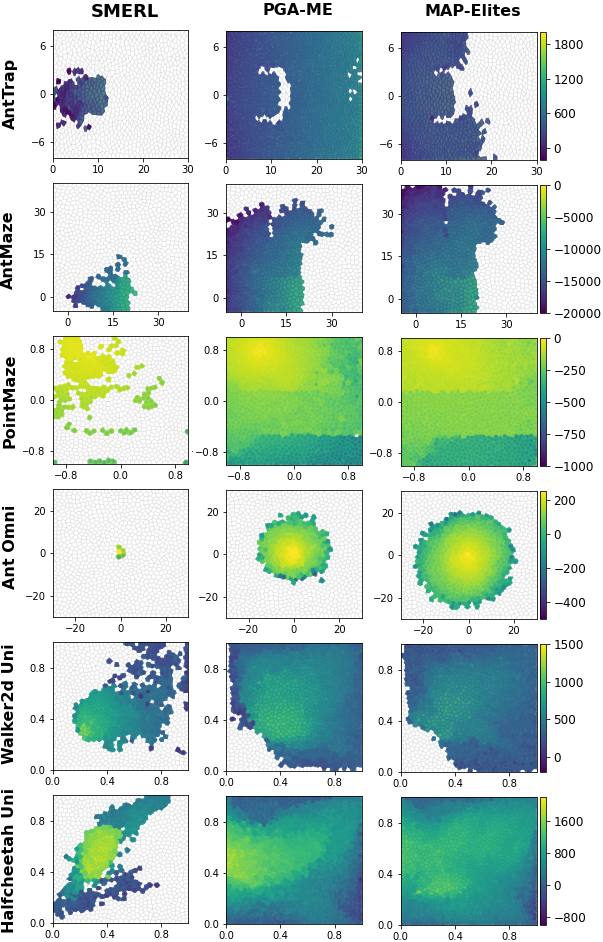}
    \caption{Final grids obtained on 6 \brax environments with \qdax implementations. For \anttrap, \antmaze, \ptmaze and \antomni, the behavior descriptor is the final {x, y} position. For \walkeruni and \hcuni, it is the proportion of contact time of the feet on the ground. Reports \smerl, \pgame and \me after two hours of training.}
    \label{fig:final_grids}
\end{figure}

To demonstrate the speed of our implementations, \autoref{table:training_step} reports the number of training steps achieved in two hours by the implementation of \qdax on a single GPU. \autoref{fig:qd_metrics_step} reports the evolution of fitness, QD score and coverage along time of the algorithms \smerl, \me and \pgame on the seven tasks mentioned above. We also report the evolution of those metrics along environments interactions on \autoref{fig:qd_metrics_step}. Finally, \autoref{fig:final_grids} shows the archives of resulting policies from the different algorithms where each cell corresponds to a policy and different behaviors.

%
%
%

\begin{table} 
    \begin{center}
    \begin{tabular}{lccccc}
        
        \toprule
        \textbf{Environment} & \me & \pgame & \smerl & \dads & \diayn \\
        \midrule
        \textbf{Ant Omni} & $1.7 \times 10^9$ & $1.85 \times 10^8$ & $1.01 \times 10^7$ & $8.9 \times 10^6$ & $1.01 \times 10^7$ \\
        \textbf{Ant Uni} & $2.45 \times 10^9$ & $1.9 \times 10^8$ & $1.0 \times 10^7$ & $7.8 \times 10^6$ & $9.97 \times 10^6$ \\
        \textbf{Ant Maze} & $1.05 \times 10^9$ & $1.7 \times 10^8$ & $7.09 \times 10^6$ & $8.62 \times 10^6$ & $9.70 \times 10^6$ \\
        \textbf{Ant Trap} & $1.08 \times 10^9$ & $1.3 \times 10^8$ & $9.81 \times 10^6$ & $8.46 \times 10^6$ & $9.73 \times 10^6$ \\
        \textbf{Halfcheetah Uni} & $9.83 \times 10^8$ & $1.39 \times 10^8$ & $1.08 \times 10^7$ & $8.43 \times 10^6$ & $1.07 \times 10^7$ \\
        \textbf{Point Maze} & $2.41 \times 10^9$ & $3.01 \times 10^8$ & $1.09 \times 10^7$ & $9.70 \times 10^6$ & $1.07 \times 10^7$ \\
        \textbf{Walker2d Uni} & $1.95 \times 10^9$ & $2.40 \times 10^8$ & $1.11 \times 10^7$ & $8.94 \times 10^6$ & $1.11 \times 10^7$ \\
        \midrule
        \textbf{Average} & $1.66 \times 10^9$ & $1.93 \times 10^8$ & $9.97 \times 10^6$ & $8.70 \times 10^6$ & $1.03 \times 10^7$ \\
        \bottomrule
    
    \end{tabular}
    \end{center}
    \caption{Number of training steps carried out during two hours of training by the various methods under study on seven \brax tasks of \qdax. Averaged over 5 seeds.}
    \label{table:training_step}
\end{table}

\begin{figure} 
    \centering
    \includegraphics[width=0.9\textwidth]{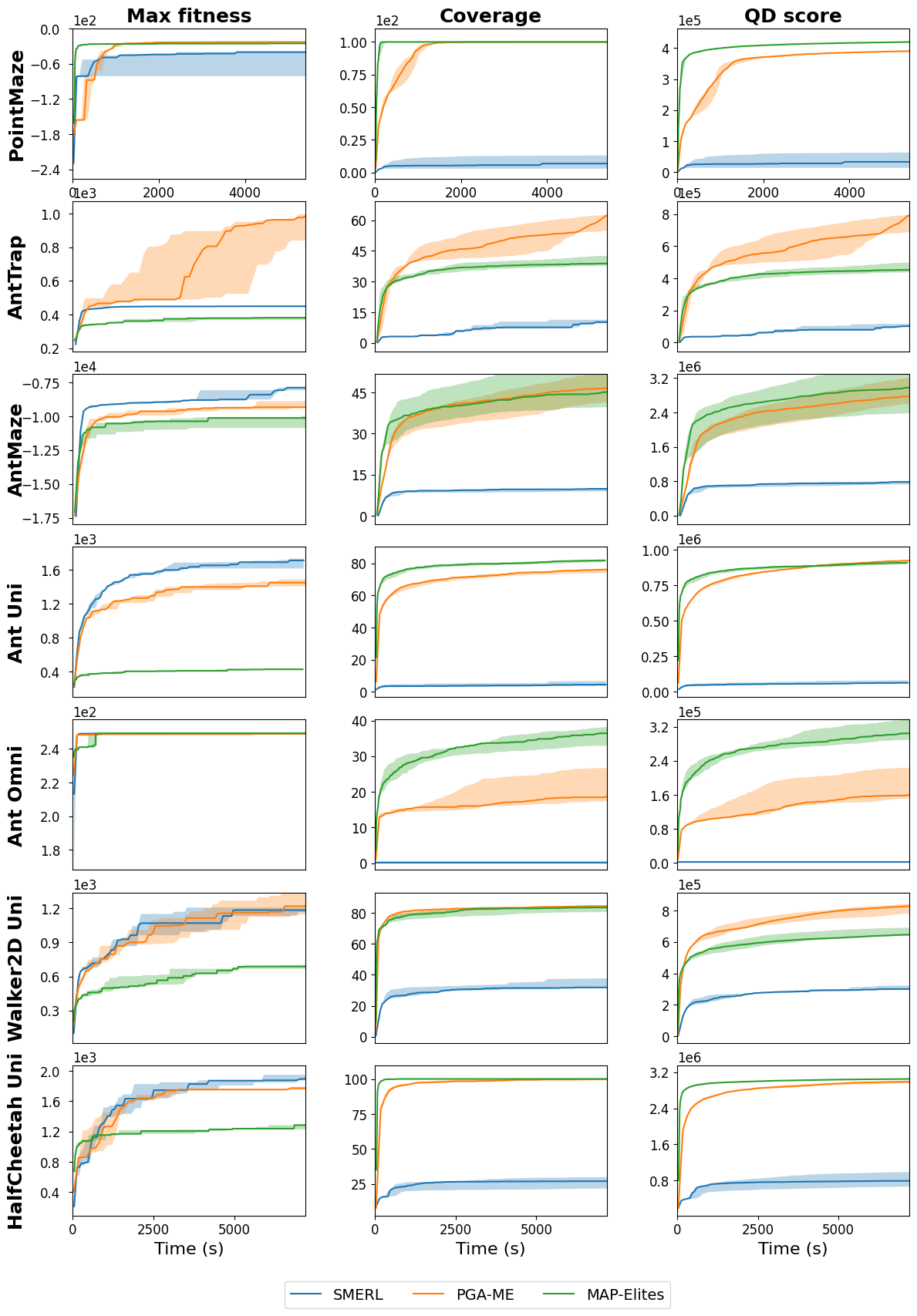}
    \caption{Evolution of the maximum fitness, coverage and \qd score along environment interactions, during a training phase. Reports algorithms \smerl, \pgame and \me  on 2 hours of training.}
    \label{fig:qd_metrics_time}
\end{figure}

\section{Fast implementations}
\label{app:fast}

\qdax is a package written in \jax, hence all the implementations can be just-in-time compiled (jit) and run on hardware accelerators like GPUs and TPUs with no code overhead. This combines simplicity and efficiency, making \qdax particularly suitable for practitioners that do not have access to large compute resources. In this section, we put ourselves in the situation of a practitioner that has a machine with an affordable GPU and wants to use existing available open-source implementations by just installing and running them.

We consider a few open-source implementations of libraries and algorithms which runtime-performance is reported in a paper, reported by its authors, or by simply running the implementation to compare the runtime-performance to the one that someone can expect by running \qdax on an affordable GPU (Quadro RTX 4000).

For \pgame, we compare \qdax implementation with the only other available open-source implementation which is the provided author implementation\footnote{\url{https://github.com/ollenilsson19/PGA-MAP-Elites}}. The author implementation utilizes PyTorch as its deep learning framework. It also uses multiprocessing over CPU devices to handle its environment evaluation and is not directly compatible with GPUs even for training the networks via gradient descent. Hence, to exploit this implementation effectively, a large number of CPUs is required, which is not an option for most practitioners. The original implementation requires approximately 36 hours to perform a run of $10^9$ steps with 36 CPUs, which corresponds to $2.8 \times 10^7$ steps an hour. Our implementation of \pgame achieves $10^8$ steps an hour. Hence, a practitioner can perform runs of \pgame in a few hours on his computer instead of several days.

For \qdpg, we compare the runtime performance of \qdax's implementation with results reported by the authors of the algorithm. The author implementation is not open-source but they report that with 1 GPU and a dozen CPUs, their implementation can do $10^8$ steps in 15 hours on \anttrap. In \qdax, \qdpg can do the same on 3 hours, hence a speed-up of a factor 5.  

For \mees, we compare our implementation with the original implementation\footnote{\url{https://github.com/uber-research/Map-Elites-Evolutionary}}. This implementation requires large computing clusters to be efficient: results reported in the paper~\citep{colas2020scaling} use 1000 CPUs, which makes it completely unusable for most practitioners. Our implementation can be run on a simple GPU with a few CPUs and still match the runtime performance. The original implementation was reported to perform $6 \times 10^8$ environment steps an hour ($3 \times 10^{10}$ in two days for the \antmaze experiment). Our implementation performs $4 \times 10^8$ steps an hour.

For \me, the reader should refer to \citet{lim2022accelerated} for a thorough analysis of the performance to be expected in \qdax.

\end{document}